\documentclass[draftclasofoot, onecolumn]{IEEEtran}
\IEEEoverridecommandlockouts
\usepackage{array}
\usepackage[caption=false]{subfig}
\usepackage[noadjust]{cite}
\usepackage[bookmarks=true,breaklinks=true,colorlinks,linkcolor=red,citecolor=blue,urlcolor=BlueViolet]{hyperref}
\usepackage{multirow}
\usepackage{amsmath,amssymb,amsfonts}

\usepackage{graphicx}
\usepackage{threeparttable}
\usepackage{textcomp}
\usepackage{xcolor}
\def\BibTeX{{\rm B\kern-.05em{\sc i\kern-.025em b}\kern-.08em
    T\kern-.1667em\lower.7ex\hbox{E}\kern-.125emX}}
\usepackage{booktabs}
\usepackage{tablefootnote}
\usepackage{float}
\usepackage{algorithmic}
\usepackage{textcomp}
\usepackage{booktabs}

\begin{document}

\title{Multi-Head Self-Attending Neural Tucker Factorization}

\author{Yikai Hou\textsuperscript{*} \quad \quad Peng Tang\textsuperscript{*}\thanks{\textsuperscript{*}College of Computer and Information Science, Southwest University, Chongqing, China (yikaih@email.swu.edu.cn, tangpengcn@swu.edu.cn)}
}

\maketitle

\begin{abstract}
Quality-of-service (QoS) data exhibit dynamic temporal patterns that are crucial for accurately predicting missing values. $\mathcal{L}$ These patterns arise from the evolving interactions between users and services, making it essential to capture the temporal dynamics inherent in such data for improved prediction performance. As the size and complexity of QoS datasets increase, existing models struggle to provide accurate predictions, highlighting the need for more flexible and dynamic methods to better capture the underlying patterns in large-scale QoS data. To address this issue, we introduce a neural network-based tensor factorization approach tailored for learning spatiotemporal representations of high-dimensional and incomplete (HDI) tensors, namely the Multi-head Self-attending Neural Tucker Factorization (MSNTucF). The model is elaborately designed for modeling intricate nonlinear spatiotemporal feature interaction patterns hidden in real world data with a two-fold idea. It first employs a neural network structure to generalize the traditional framework of Tucker factorization and then proposes to leverage a multi-head self-attending module to enforce nonlinear latent interaction learning. In empirical studies on two dynamic QoS datasets from real applications, the proposed MSNTucF model demonstrates superior performance compared to state-of-the-art benchmark models in estimating missing observations. This highlights its ability to learn non-linear spatiotemporal representations of HDI tensors.
\end{abstract}


\section{Introduction}

With the rapid development of modern service-oriented technologies, such as cloud computing \cite{XiaY15,HM17}, edge computing \cite{HuaH23}, and the increasing adoption of Internet of Things (IoT) systems \cite{ZhangX24}, the number of available web services is growing at an unprecedented rate. This expansion has led to a proliferation of service offerings with similar functionalities, creating a challenge for users to effectively identify and select the most suitable services for their needs \cite{WangJ24}. In this context, Quality of Service (QoS) plays a crucial role in service selection, encompassing factors such as response time, throughput, reliability, and cost. From the perspective of both service providers and users, evaluating and predicting QoS is significant to ensuring optimal service delivery and user satisfaction \cite{LiuY04, DengS16,BiF22,XiaY13,LiM21+}. 

As a result, QoS prediction methods, which aim to infer missing or unknown QoS values, have emerged as an essential tool to facilitate service selection and decision-making processes \cite{WuD22, YeF21, PengZ22,LuoX19_,ChenM24+,LuoX14+,SunX18}. Among the various techniques proposed for QoS prediction, Latent Factor Analysis (LFA)-based models have gained significant attention \cite{LuoX21, LuoX16, WuD23, LuoX18, LuoX22_,LuoX18_,LuoX21__,WuD22_,ChenJ23}. These methods focus on learning a low-dimensional latent feature space by decomposing a user-service QoS matrix, where rows represent users, columns represent services, and matrix entries correspond to observed QoS values. The missing entries in this matrix can then be predicted by the interactions between latent features of users and services, typically through the inner product of these latent vectors. LFA-based approaches have  been proven effective in predicting QoS across various scenarios due to their strong ability to generalize from sparse data \cite{LuoX19,LuoX14, TangM16,LuoX21+,LuoX22+,LiJ24,ZhongY24,LiuZ18,WangQ22,YuanY24, LiW22, LuoX21++}.

However, a key limitation of many LFA-based models is their static nature—they typically assume that the QoS values do not change over time. In reality, the QoS experienced by users can fluctuate due to various factors, such as network conditions, server load, and user behavior. This temporal variability is a significant challenge for QoS prediction models, as ignoring it leads to suboptimal predictions. To address this, researchers have extended traditional LFA-based methods to incorporate temporal dynamics, resulting in Tensor Factorization (TF)-based models. These models represent QoS data as 3D tensors, where the third dimension corresponds to time, allowing for the modeling of temporal patterns in the data.

For example, latent Factorization of Tensors (LFT) approaches \cite{LuoX22, LuoX23,LuoX20,ZhangW14,WuH22,WuH22_,TangP24_, LuoX21_} have been developed. A biased non-negative CP-based LFT approach \cite{LuoX20} has demonstrated its effectiveness and efficiency on QoS prediction via capturing the temporal dynamics in data. Density-oriented principles such as single latent factor-dependent nonnegative and multiplicative updates on tensors (SLF-NMUT) \cite{LuoX22} and alternating direction method-based sequential task learning (ADM-STL) \cite{WuH22_}  have been proposed to handle data sparsity effectively  and enhance robustness. Those studies essentially adopts multi-linear approaches and overlooks the nonlinear pattern hidden in spatiotemporal data. Inspired by the superior nonlinear representation learning ability of neural networks, neural network (NN)-based LFT approaches \cite{WuH21} are proposed to characterize the complex nonlinear dependency across different spatiotemporal dimensions.

Despite these advancements, predicting QoS with high accuracy remains a challenging problem, particularly as the size and complexity of web service datasets continue to grow. Effective prediction requires a model that can handle both the sparsity and temporal dynamics of QoS data. To solve these problems, this paper proposes the MSNTucF model and aims to make the following contributions: 1) an effective multi-layer neural network-based Tucker factorization model for accurate spatiotemporal representation learning; 2) a multi-head self-attending latent interaction learning module for capturing the complex nonlinear relationships among different dimensions in QoS data.

\begin{figure*}[t]
\centering
\includegraphics[width=18cm]{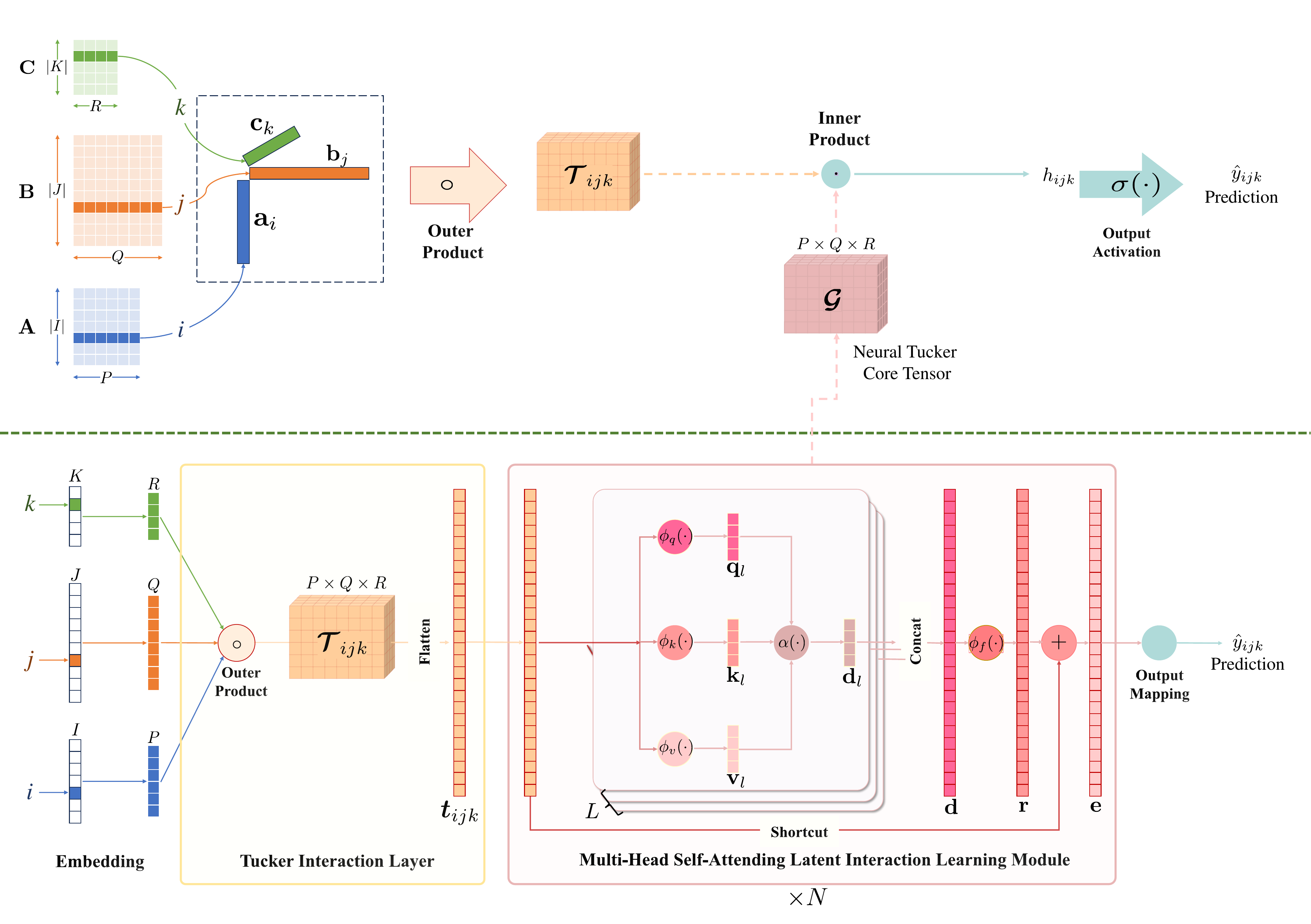}
	\caption{The Multi-Head Self-Attending Neural Tucker Factorization. Colors indicate the equivalence and dashed lines denote the inference from the operations in the latent interaction learning module.}
	\label{fig:model}
\end{figure*}
\section{Preliminaries}\label{Pri}
\subsection{ Latent factorization of tensors}
Latent factor analysis (LFA) and latent factorization of tensors (LFT) are closely related techniques used to uncover hidden patterns in complex, high-dimensional data. LFA typically involves the decomposition of observed variables into a set of unobserved latent factors, which explain the underlying structure of the data \cite{ShiX20,YuanY20,ChenD21,WuD21+,WuD21++,WangJ24+,WuD23+,YuanY24+,LiW23,LiW22+,YuanY23+,WuH21++}. This is often achieved through methods like factor analysis, where observed variables are modeled as linear combinations of latent factors, allowing for dimensionality reduction and the identification of key drivers behind the data. In this approach, the observed data matrix is approximated by a product of matrices representing the latent factors and their corresponding loadings.

Building on this concept, latent factorization of tensors extends LFA to higher-order data structures, such as multidimensional tensors\cite{ChenM21+, ChenM24++,WuH24+,ZhongY23,MiJ23}. A tensor is a multidimensional array, and its dimensionality is referred to the rank or mode. Let $\boldsymbol{\mathcal{X}} \in\ \mathbb{R}^{I_1\times I_2 \ldots I_n}$ be an input tensor, where $\mathcal{X}$ has N modes, with each mode's dimensionality denoted by $I_1$ through $I_n$. A common approach to tensor factorization involves representing the original tensor as a combination of latent factor matrices, which provide a compact yet expressive representation of the data. Such decompositions reduce the complexity of the tensor while preserving the essential structures and dependencies across different dimensions. The core tensor, often of reduced rank, serves as a bridge that encodes the multi-way interactions among the latent factors across different modes. Depending on the specific decomposition method, such as Canonical Polyadic (CP) decomposition, Tucker decomposition, or Tensor Train (TT) decomposition, the structure of the core tensor and latent factors may vary.

One of the key advantages of latent factor tensorization (LFT) is its ability to efficiently model high-dimensional and complex data while maintaining the interpretability of the learned latent factors. Unlike traditional matrix decomposition methods, tensor-based approaches explicitly account for multi-way interactions, making them particularly useful in applications such as recommender systems, spatiotemporal modeling, and scientific data analysis. By leveraging the multi-linear structure inherent in tensors, LFT not only enhances representation learning but also facilitates efficient computation and storage, making it a powerful tool for analyzing structured, high-dimensional data.

\subsection{Tucker decomposition}
Consider a Mode-N HDI tensor $\boldsymbol{\mathcal Y}$, where the set of observed elements is denoted by  $\Lambda$, with an unknown element set $\Gamma$ ($|\Gamma|\gg|\Lambda|$). To estimate the missing entries, the LFT model generates a low-rank approximation $\boldsymbol{\hat{\mathcal{Y}}}$. This is achieved by modeling entity interactions, where rank-one tensors are constructed based on specific feature interaction schemes defined in various tensor decomposition frameworks. Tucker decomposition \cite{Jang2020,Kim2007,MGiraud2023,MLi2024,TKim2022,YQiu2019,KFonał2019,JYin2020,JYao2024,LSun2022,HXiang2023,YQiu2022,TLiu2023,SAhmadi2021,CQian2021,SXu2024,JFeng2022,YHan2022,XTong2023,LLi2022}, in particular, accounts for all possible feature interactions and introduces a core tensor that assigns weights to the rank-one tensors for the approximation. In this paper, we focus on the mode-3 case, which is expressed as
\begin{equation}
    \boldsymbol{\hat{\mathcal{Y}}} = \sum_{p=1}^{P}\sum_{q=1}^{Q}\sum_{r=1}^{R} g_{pqr}*\boldsymbol{\mathcal{A}_{pqr}},
\end{equation}
where $g_{pqr}$ represents the $(p,q,r)$-th element of the core tensor $\boldsymbol{\mathcal{G}}$, quantifying the strength of feature interactions. $\boldsymbol{\mathcal{A}_{pqr}}$ denotes the rank-one entity tensor, constructed as the outer product of the latent feature vectors $a_p \in \mathbb{R}^I$, $b_q \in \mathbb{R}^J$ and $c_r \in \mathbb{R}^K$ along the three modes.

\subsection{Neural Tucker Factorization}
Neural Tucker Factorization (NeuTucF)\cite{TangP24} leverages neural networks to implement a density-oriented approach, enabling element-wise approximations from a latent interaction perspective in Tucker decomposition. This can be expressed as
\begin{equation}
    \hat{y}_{ijk} = \boldsymbol{\mathcal{G}}\cdot \boldsymbol{\mathcal{T}}_{ijk} = \sum \boldsymbol{\mathcal{G}}\odot \boldsymbol{\mathcal{T}}_{ijk} =  \sum_{p=1}^{P}\sum_{q=1}^{Q}\sum_{r=1}^{R} g_{pqr}t_{pqr}^{(ijk)},
\end{equation}
where $\cdot$ and $\odot$ denote inner product and Hadamard product operation between two tensor respectively. $\boldsymbol{\mathcal{T}}_{ijk}$ is the rank-one interaction tensor constructed as the outer product of the latent feature vectors $a_i \in \mathbb{R}^P$, $b_j \in \mathbb{R}^Q$ and $c_k \in \mathbb{R}^R$ along the three modes. 

\section{Multi-head Self-Attending Neural Tucker Factorization}
\subsection{Objective of MSNTucF}
To precisely predict high dimensional and incomplete tensor, we propose the MSNTucF model, which leverages the core concept of NeuTucF and enhances it with nonlinearity by integrating a multi-head self-attending module for latent interaction learning. Fig. \ref{fig:model} shows the basic structure of our MSNTucF model, where $\phi(\cdot)$ represents linear transformation and $\alpha(\cdot)$ denotes the process of calculating self-attention. First, the inputs to the model are a set of triples $(i, j, k)$ , where $i, j ,k$ represent the indexes of the different tensor modes of one valid piece of data. After that, one hot encoding process will transform each element of these triples into binarized sparse vectors. Then, this sparse vectors are mapped into a dense embedding vector by a embedding layer, so we get three mode embeddings $\mathbf {a}_i$, $\mathbf{b}_j$, and $\mathbf{c}_k$.

Subsequently, the three mode embeddings are introduced to a Tucker Interaction layer. This layer can capture latent interaction among spatiotemporal features of different dimensions. Compared to simple feature concatenation, the outer product can provide more complete interactions for individual dimensional features. The spatiotemporal interaction tensor $\boldsymbol{\mathcal{T}}_{ijk}$ can be defined as the outer product of the embedding vector $\mathbf{a_i}$, $\mathbf{b_j}$, and $\mathbf{c_k}$ in the form of
\begin{equation}
    {\mathbf{\mathcal{T}} _{ijk}} = \mathbf{a}_i \circ \mathbf{b}_j \circ \mathbf{c}_k,
\end{equation}
where $ \circ $ denotes the outer product operator and the size of  $\boldsymbol{\mathcal{T}}_{ijk}$ is $P \times Q \times R$. Each component of the tensor corresponds to an interaction involving the latent factor ${a_{ip}}$, ${b_{jq}}$, and ${c_{kr}}$. The interaction tensor will be transformed into a vector by a flattening operation, which facilitates subsequent input to the neural network structure. The process can be expressed as
\begin{equation}
    {\mathbf{t}_{ijk}} = \theta ({\mathcal{T}_{ijk}}),
\end{equation}
where $\theta $ denotes the flatten operation.
                                            
Next, We utilize the multi-head self-attention mechanism\cite{Vaswani17} to learn the spatiotemporal interactions in $\mathbf{t}_{ijk}$ . From the multi-head self-attending module we can get an output denoted as $\mathbf{e}_{ijk}$ and  this module will be looped N times to get a final result $\mathbf{e}_{ijk}^{(n)}$. This module can be used as a key component in MSNTucF architecture, which we will further elaborate in the next part of this section.

Finally, The result $\mathbf{o}_{ijk}^{(n)}$ passes through the linear layer to obtain the final output of the model. This paper chooses sigmoid activation as the output mapping which can be formalized as
\begin{equation}{
y}_{ijk} = \sigma (\mathbf{W}\mathbf{e} _{ijk}^{(n)}),\end{equation}
where $\sigma $ represents the sigmoid function and $\mathbf{W} \in \mathbb{R}^ {1 \times d_{\text{model}} }$ denotes the weight matrix of the linear layer. $d_\text{model}$ is the length of each interactive sequence and the final output sequence obtained by the model.

Multi-Head Self-Attending Module:  The module is used to learn the potential and latent nonlinear relationships in the spatiotemporal interaction $
\mathbf{t}_{ijk}$ and mine them further.

a) Linear Transformation to generate Queries, Keys and Values. The input to this module is a sequence $\mathbf{t} \in \mathbb{R}^{ d_{\text{model}}}$. In order to adjusting attention weights based on feature interactions, we create separate matrices for queries q, keys k, and values v by applying linear transformations without bias as weight matrices for the input interactive sequence. This allows the model to adaptively focus on important information, enhancing performance while mitigating the effects of noise or irrelevant data. The process can be formulated as
\begin{equation}\mathbf{q}_l = {\mathbf{W}^Q_l}\mathbf{t},{\kern 1pt} {\kern 1pt} {\kern 1pt} {\kern 1pt} {\kern 1pt} {\kern 1pt} {\kern 1pt} {\kern 1pt} \mathbf{k}_l =  {\mathbf{W}^K_l}\mathbf{t},{\kern 1pt} {\kern 1pt} {\kern 1pt} {\kern 1pt} {\kern 1pt} {\kern 1pt} {\kern 1pt} {\kern 1pt} \mathbf{v}_l =  {\mathbf{W}^V_l}\mathbf{t},\end{equation}
where ${\mathbf{W}^Q_i} \in \mathbb{R}^{d_k \times  d_{\text{model}}},{\mathbf{W}^K_i}\in \mathbb{R}^{d_k \times  d_{\text{model}}}, {\mathbf{W}^V_i}\in \mathbb{R}^{d_k \times  d_{\text{model}}}$ are learned weight matrices and $d_{k}$ is the length of each $\mathbf{q}$, $\mathbf{k}$ and $\mathbf{v}$. Through $L$ identical operations, we can obtain $L$ $(\mathbf{q},\mathbf{k},\mathbf{v})$ triples, $L$ being the number of attention head. Generally, we set $d_{\text{model}} = d_k\times L$.

b) Scaled Outer Product Attention. The similarity between the interacting queries and the keys is calculated by dot product to determine which positions have stronger correlation with each other during the computation of serial attention. However, we deal with a single vector and we are concerned with the similarity between each element of the vector. Instead of dot product, we choose outer product to get the correlation between each position in the vector. Each attention head calculates the outer product of query and key, scales the result, and then applies a softmax function to obtain the attention scores. Here, we use a scaling factor $\sqrt{d_k}$ to avoid large outer product values caused by high-dimensional features, which can lead to unstable attention scores, and then softmax is used to convert the computed attention scores into a probability distribution. The calculation of attention score can be expressed as

\begin{equation}\mathbf{d}_{l} = \text{Softmax}(\frac{{\mathbf{q}_l \circ \mathbf{k}_l}}{{\sqrt {{d_k}} }}) \cdot \mathbf{v}_{l} .\end{equation}
Then, attention score is multiplied by $\mathbf{v}_l$ in order extract the most important information from the value vector by weighted summation. This step of the operation enables the model to intelligently focus on the most relevant information for effective interaction fusion. Each attention head generates a sequence of interactions $\mathbf{d}_{l}$.

c) Concatenation of Heads. Output interactive sequence $\mathbf{d}_{l}$ from all heads are concatenated to a complete sequence. The connection process is as follows
\begin{equation}\mathbf{r}  = \text{Linear}\left( {\text{Concat}(\mathbf{d}_{1},\mathbf{d}_{2},...,\mathbf{d}_{L}}) \right).\end{equation}
Through linear transformations, the model is able to fuse information from the outputs of multiple attention heads, learn how to better extract and combine information from different perspectives, and enhance the expressive power of the model.  

d) Residual and Layer-norm. A layer-norm connection around each of the layer and a residual module are employed here. That is
\begin{equation}
    {\mathbf{e}_{ijk}^{(i)}} = \text{Layernorm}(\mathbf{e}_{ijk}^{(i-1)} + {\mathbf{r}}),
\end{equation}
where $i$ represents the number of times the module is currently looping, and $i-1$ is the result obtained in the previous round of looping. Here, We integrate the residual mechanism to allow the self-attending module to preserve the original information of the input while fully utilizing the contextual information, so ensure that the model can always maintain a portion of the input information that is not weakened or lost in multiple transformations. Besides, the weight parameters of different layers may have different scales, leading to large differences in the distribution of the outputs of each layer, which can make the training of subsequent layers difficult. Layer normalization normalizes the inputs of each layer, making their distribution more consistent across layers. In this way, the network is able to maintain a stable delivery of input information at each layer, which improves the training effect and convergence speed of the model.

\subsection{Learning Schem}
 Euclidean distance\cite{Dokmanic15} is a direct measure of the gap between predicted and true values, it calculates the difference between predicted and actual observations. To train the model and optimize its parameters, the loss function is defined as a measure of the model's capability to approximate the QoS data. This is achieved by computing the Euclidean distance between the original value,  $\boldsymbol{\mathcal{Y}}$, and its approximation,  $\boldsymbol{\mathcal{\hat Y}}$, as outlined in this paper. Anyway, our model performs strongly with incomplete data, and the majority of some datasets entries are unknown, it suffices to focus on the information derived from the known element set $\Lambda$. Consequently, this leads to the following outcomes
\begin{align}
\mathcal{E}&=\frac{1}{2}\sum_{y_{ijk} \in \Lambda}(y_{ijk}-\hat{y}_{ijk})^2 \nonumber\\
&=\frac{1}{2}\sum_{y_{ijk} \in \Lambda}\Biggl(y_{ijk}-  \sigma (\mathbf{W}\mathbf{e} _{ijk}^{(n)})\Biggr)^2.
\end{align}
Its optimization can be achieved by using stochastic gradient descent (SGD) \cite{LuoX21____} or its variants, such as AdaGrad, Adam, RMSProp, and others. These optimizers are highly universal and help in tuning the model parameters to minimize the loss function. However, Adam is ideal in our task due to its ability to better balance training speed and accuracy when dealing with spatiotemporal data, especially data containing a large number of missing values or noise. Additionally, When calculating the attention scores, we applied dropout to randomly discard some of the weights, which helps to prevent the model from being overly reliant on some specific locations.

\newcolumntype{C}[1]{>{\centering\arraybackslash}m{#1}}
\begingroup
\begin{table}[]
\renewcommand\arraystretch{1}
\scriptsize
\centering
\caption{A Summary of WSDREAM Data}
\label{tab:data}
\renewcommand\arraystretch{1.5}
\begin{tabular}{C{2cm}C{3cm}C{3cm}}
\noalign{\hrule height 1pt}
\textbf{Dataset} & \textbf{QoS-RT} & \textbf{QoS-TP} \\
\hline
Datatype                   & Response time      & Throughput        \\
Value Scale               & 0-20 s       & 0-1000 kbps        \\
No. of Users               & 142       & 142        \\
No. of Services               & 4500       & 4500        \\
No. of Time slices               & 64       & 64        \\
No. of Records               & 30,287,611       & 30,287,611        \\

\noalign{\hrule height 1pt}
\end{tabular}
\end{table}
\endgroup

\begingroup
\renewcommand\arraystretch{1}
\newcolumntype{D}[1]{>{\centering\arraybackslash}m{#1}}
\begingroup
\begin{table}[]
\renewcommand\arraystretch{1}
\scriptsize
\centering
\caption{Details of Experiment Datasets}
\label{tab:datas}
\renewcommand\arraystretch{1.5}
\begin{tabular}{C{1,8cm}C{1.8cm}C{2cm}C{2.0cm}C{2cm}}
\noalign{\hrule height 1pt}

\textbf{Dataset} & \textbf{No.} & \textbf{Train:Valid:Test} & \textbf{Density (\%)} \\
\hline
\multirow{2}{*}{QoS-RT}  & \textbf{D1} &          2:6:92        & 1.329        \\
   & \textbf{D2}                   & 5:15:80          & 3.323        \\
\hline
\multirow{2}{*}{QoS-TH}  &\textbf{D3}        & 2:6:92          & 1.255        \\
   & \textbf{D4}                    & 5:15:80         & 3.136        \\
\noalign{\hrule height 1pt}
\end{tabular}
\end{table}
\endgroup

\begingroup
\renewcommand\arraystretch{1}
\begin{table*}[t]
\centering
\caption{The Summary of Results}
\label{tab:result}
\scriptsize
\renewcommand\arraystretch{1.5}
\begin{tabular}{C{0.3cm}
C{0.6cm}C{0.6cm}C{0.6cm}C{0.01cm}
C{0.6cm}C{0.6cm}C{0.6cm}C{0.01cm}
C{0.6cm}C{0.6cm}C{0.7cm}C{0.01cm}
C{0.6cm}C{0.6cm}C{0.7cm}
C{1cm}C{0.6cm}C{0.8cm}}
\noalign{\hrule height 1pt}
\multirow{2}{*}{} 
& \multicolumn{3}{c}{{\textbf{D1}}} & 
& \multicolumn{3}{c}{{\textbf{D2}}} & 
& \multicolumn{3}{c}{{\textbf{D3}}} & 
& \multicolumn{3}{c}{{\textbf{D4}}}  
& \multirow{2}{*}{{\textbf{Win/Loss}}} & \multirow{2}{*}{{\textbf{Rank}}}  & \multirow{2}{*}{{\textbf{p-value}}} \\
\cline{2-4} \cline{6-8} \cline{10-12} \cline{14-16} 
 & {\textbf{MAE}}    & {\textbf{MRE}}    & {\textbf{RMSE}}  &   
 & {\textbf{MAE}}    & {\textbf{MRE}}    & {\textbf{RMSE}}  & 
 & {\textbf{MAE}}    & {\textbf{MRE}}    & {\textbf{RMSE}}  &    
 & {\textbf{MAE}}    & {\textbf{MRE}}    & {\textbf{RMSE}}  
 &                          \\
\hline
{\textbf{M1}}                     
& 0.6942 &	1.3181  &	1.9361  & 
&	0.6785  &	1.1595  &	1.9035  & 
&	4.8487  &	0.9053  & 35.6790  & 
& 4.7416  &	0.8737  &	35.1453 
& \textbf{0/12 }  
& \textbf{4.50 }
& {\textbf{4.88E-4}}\\
{\textbf{M2}}                     
& 0.6916  & 1.2943  & 1.9331    & 
& 0.6746  & 1.1442  & 1.8938  &  
& 4.7500  & 0.9848 & 36,3399    & 
& 4.6664  & 0.9302  & 35.0317 
& \textbf{0/12}
& \textbf{4.08 }
& {\textbf{4.88E-4}}\\
{\textbf{M3}}                     
& 0.6840  & 1.4006 & 1.9299  &
& 0.6655  & 1.1939  & 1.8959  &
& 4.6440  & 0.8926  &35.9957  &
& 4.5346  & 0.8449  & 35.4498 
& \textbf{0/12 }
& \textbf{3.75 }
& {\textbf{4.88E-4}}\\
{\textbf{M4}}                     
& 0.7202  & 1.3750 & 2.0835  &
& 0.6846  & 1.1214  & 2.0683  &
& 4.9304  & 0.9184  & 36.1126  &
& 4.6678  & 0.8723  & 35.0856 
& \textbf{0/12 }
& \textbf{5.17}
& {\textbf{4.88E-4}}\\
{\textbf{M5}}                    
& 0.7342  & 1.3770  & 1.9599  &
& 0.7235 & 1.5212  & 1.9388  &
& 5.6908  & 0.9136 & 39.6473  &
& 5.4213  & 0.9566  & 39.0000 
& \textbf{0/12 }
& \textbf{6.50} 
& {\textbf{4.88E-4}}\\
{\textbf{M6}} 
&0.6804  & 1.3210  & 1.8770   &
&0.6630  & 1.1360  & 1.8351   &
& 4.8799  & 0.9159  & 35.6002   &
& 4.6946  & 0.8433  & 34.7706 
& \textbf{0/12}
& \textbf{3.00}
& {\textbf{4.88E-4}}\\
{\textbf{M7}}                    
& \textbf{0.6607} & \textbf{1.2281}  & \textbf{1.8136}   & 
& \textbf{0.6158} & \textbf{1.0592}  & \textbf{1.7330}  &
& \textbf{4.6398}  & \textbf{0.8421}  & \textbf{34.0577}   &
& \textbf{4.2504} & \textbf{0.8111} & \textbf{31.1317}
& \textbf{12/12}  
& \textbf{1.00 }
& \textbf{-}\\
\noalign{\hrule height 1pt}    
\end{tabular}
\end{table*}
\endgroup
\section{Experiments}
\subsection{Experiment Setting}
To fully prove the prediction performance and generalization ability of the proposed model, we conduct extensive experiments on two real-world datasets. Four datasets, D1 through D4, are extracted from WSDream with varying split ratios for evaluation purposes. The basic information about the data is organized in Table \ref{tab:data} and details of the experiment datasets is provided in Table \ref{tab:datas}. The data distribution is heavily skewed, characterized by substantial variances, and does not align with the probabilistic presuppositions inherent in low-rank factorization models. So, we preprocess data by log transformation and min-max normalization in all the experiments to make data more normal distribution-like to fit that assumption.

The experiments were performed on a platform with a 2.50-GHz 13th Gen Intel(R) Core(TM) i5-13400F CPU and one NVIDIA GeForce RTX3050 GPU with 32-GB RAM. To assess the model's performance in predicting unknown entries of the HDI tensors, this paper employs three evaluation metrics: Mean Absolute Error (MAE), Mean Relative Error (MRE) and Root Mean Square Error (RMSE). Together, these metrics holistically assess model performance—MAE ensures operational stability in resource-sensitive contexts, MRE enables cross-metric comparisons, and RMSE highlights reliability under extreme network conditions, collectively addressing the challenges of sparse tensor completion in  prediction. They can be expressed as
\begin{align}
\text{MAE}& = \frac{1}{|\Omega_{\text{test}}|} \sum_{(i,j,k) \in \Omega_{\text{test}}} \left| y_{ijk} - \hat{y}_{ijk} \right|, \\
\text{MRE} &= \frac{1}{|\Omega_{\text{test}}|} \sum_{(i,j,k) \in \Omega_{\text{test}}} \frac{\left| y_{ijk} - \hat{y}_{ijk} \right|}{|y_{ijk}|}, \quad \\
\text{RMSE} &= \sqrt{ \frac{1}{|\Omega_{\text{test}}|} \sum_{(i,j,k) \in \Omega_{\text{test}}} \left( y_{ijk} - \hat{y}_{ijk} \right)^2 }.
\end{align}

Our proposed MSNTucF model is compared with several state-of-the-art models which are capable of delivering precise forecasts for missing data within high-dimensional  and incomplete (HDI) tensors. The compared models include: (a) M1: NNCP \cite{ZhangW14}; (b) M2: CTF \cite{YeF21}; (c) M3: BNLFT \cite{LuoX20}; (d) M4: BTTF \cite{ChenX22};  (e) M5: Neural Collaborative Filtering \cite{HeX17}; (f) M6: NeuTucF \cite{TangP24}; (g) M7: Our MSNTucF model. All the model are implemented with Python 3.10.12 and Pytorch 2.4.1.

To guarantee an equitable assessment, the rank for models M1 through M4 has been standardized to a value of 5, and for models M5 to M7, the embedding dimension for each input has been uniformly set to 5. Additionally, the number of attention heads for our models has been fixed at 25, with the recurrence count set to 4. The remaining hyperparameters for each model have been meticulously optimized through a grid search approach to secure the best possible outcomes. To mitigate the impact of stochastic parameter initialization, each model has been executed on ten separate times, and the average performance metrics have been computed.

\begin{figure}[t]
\centering
\subfloat[]
{\label{fig:resulta}
\includegraphics[width=8.5cm]
{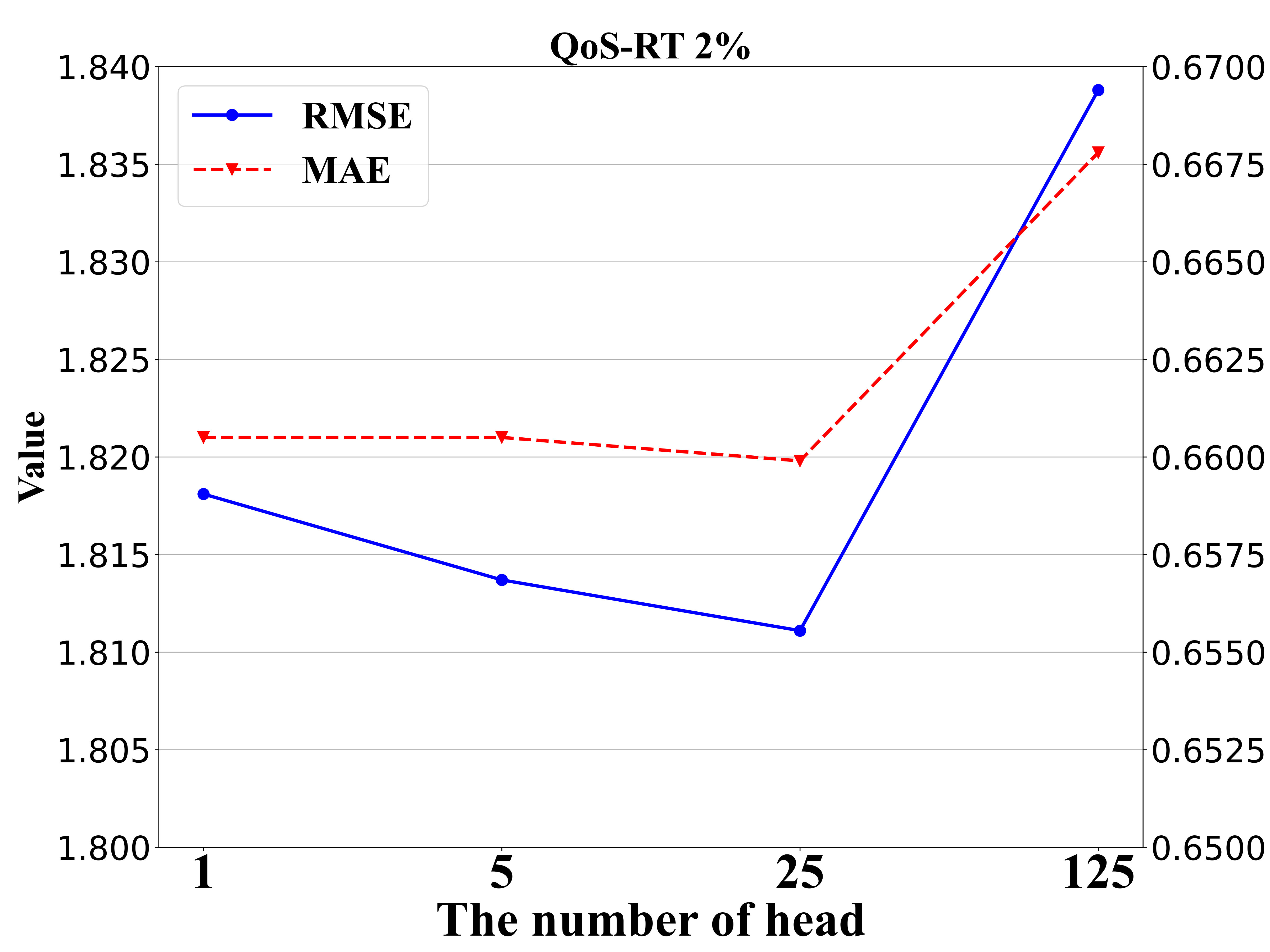} }
\quad
\subfloat[]
{\label{fig:resultb}
\includegraphics[width=8.5cm]
{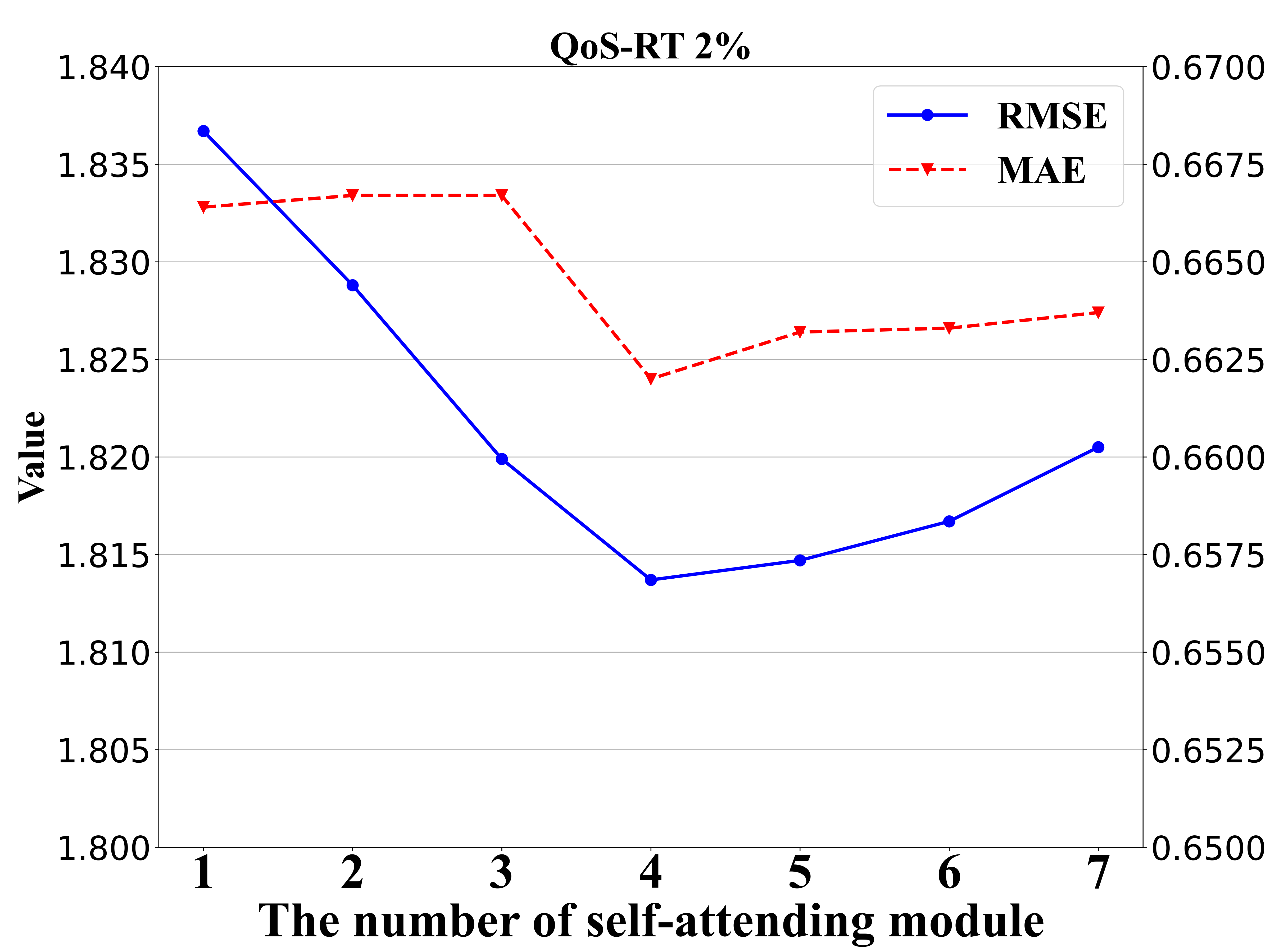}} \\
\caption{The effect of the number of head and self-attending module on performance of MSNTucF.
}
\label{fig:result}
\end{figure}

\subsection{Result Analysis}
The experimental outcomes, along with the statistical win/loss tallies, Friedman rank, and Wilcoxon test p-value are encapsulated in Table \ref{tab:result}. The key findings from these results are as follows:

1) \textit{Our MSNTucF model exhibits strong  performance  in handling  higher-dimensional and incomplete(HDI) data.} As shown in Table \ref{tab:result}, our model performs best in processing the four HDI data, where the models include those for low-rank tensor complementation M1-M4 as well as high-performance neural network models M5-M6. For instance, on D2, M7 has 8.95\%, \%, 8.64\%, 8.59\%, 16.36\%, 10.76\% and 5.56\% accuracy gain over M1 to M6 respectively in RMSE and it also has 11.42\%, 11.13\%, 12.18\%, 11.26\%, 20.17\% and 10.46\% accuracy gain over M1 to M6 respectively in RMSE on D4. This implies that MSNTucF is also a effective tensor completion model.  Furthermore, the p-values derived from the paired Wilcoxon signed rank tests are below the 0.05 threshold, signifying that there is a statistically significant difference in the representative performance between M7 and each of the preceding models, M1 through M6.

2) \textit{The number of heads with multiple self-attention have a strong effect on the experimental results.} From Fig. \ref{fig:resulta} ,the model performs better when the number of heads is set to 5 and 25. This is because, by preserving the corresponding coordinate information when flattening the tensor $\mathcal{T}_{ijk}$, we divide this vector $\mathbf{x}_{ijk}$ sequentially into 5 heads or 25 heads, the sequence corresponding to each head represents the flattened vector of a complete slice or fiber of the tensor. This indicates that a tensor can be divided into multiple slices or fibers along a specific dimension to learn the correlations within each part independently. These parts can then be integrated for effective analysis and processing. This approach provides an efficient method for QoS prediction, which can also be applied to other universal tensor representation learning models.

3) \textit{Performing the appropriate number of serial loops on our multi-head self-attending module can further enable accuracy gain.} Fig. \ref{fig:resultb} illustrate the effect of the number of self-attending module on the performance of MSNTucF. RMSE result with 4 cycles are 1.25\% better than 1 cycle and 0.3\% than 7 cycles. This result suggests that when the number of module is relatively small, the model has limited representation capability and cannot effectively capture the complex relationships of input features as well as sufficient global features. When the number of module gradually increases, the model has accumulated a certain number of local features and starts to combine global context information to gradually establish multi-scale dependencies. 4 modules happen to be the optimal point where global feature modeling is combined with local features to produce better results. When the number of module increases to higher levels, the model over fits the training data, which cannot significantly improve the performance, and will even increase the computational overhead and noise.

\section{Conclusions and future works}\label{con}

In this paper, we introduce an innovative framework, the low-rank completion and multi-head self-attention fusion network (MSNTucF), aimed at achieving more accurate QoS predictions with temporal pattern awareness. Through the integration of a multi-head self-attending module, the model successfully uncovers complex spatiotemporal dependencies generated by the Tucker interaction layer, enabling a deeper understanding of nonlinear correlations within QoS data. This innovative design highlights the potential of combining tensor-based methods with advanced neural network architectures to tackle challenges in spatiotemporal prediction tasks. Besides, based on the excellent performance of our model in QoS data, it can effectively be applied to help people do decision making and service selection.

Nevertheless, our current model is not precise enough for temporal features. To address this, we plan to integrate recurrent neural network architectures, such as LSTM, which excel at modeling temporal dependencies. In future work, we aim to explore advanced techniques for chronological dependency learning and expand the model's applicability to a broader range of real-world challenges, ultimately delivering more robust and reliable predictions.

\end{document}